\documentclass[conference]{IEEEtran}
\IEEEoverridecommandlockouts
\usepackage{cite}
\usepackage{amsmath,amssymb,amsfonts}
\usepackage{algorithmic}
\usepackage{graphicx}
\usepackage{textcomp}
\usepackage{xcolor}

\usepackage{hyperref}       
\usepackage{url}            
\usepackage{tabularx,booktabs}       
\usepackage{nicefrac}       
\usepackage{microtype}      

\usepackage{amsmath,amsbsy,amsfonts,amssymb,amsthm,units,bm}
\usepackage[mathscr]{eucal}
\usepackage{algorithm, algorithmic}
\usepackage{graphicx}
\usepackage{tikz}
\usepackage{stackengine,enumitem}
\usepackage{subfigure}
\usepackage{mathtools}
\usepackage{multirow}
\usepackage{subfiles}
\usepackage{pifont}
\usepackage{makecell}

\newcolumntype{C}[1]{>{\centering\let\newline\\\arraybackslash\hspace{0pt}}m{#1}}

\newtheorem{theorem}{Theorem}[section]

\theoremstyle{definition}

\newtheorem{definition}{Definition}[section]

\makeatletter
\newsavebox\myboxA
\newsavebox\myboxB
\newlength\mylenA

\newcommand*\xbar[2][0.75]{%
    \sbox{\myboxA}{$\m@th#2$}%
    \setbox\myboxB\null
    \ht\myboxB=\ht\myboxA%
    \dp\myboxB=\dp\myboxA%
    \wd\myboxB=#1\wd\myboxA
    \sbox\myboxB{$\m@th\overline{\copy\myboxB}$}
    \setlength\mylenA{\the\wd\myboxA}
    \addtolength\mylenA{-\the\wd\myboxB}%
    \ifdim\wd\myboxB<\wd\myboxA%
       \rlap{\hskip 0.5\mylenA\usebox\myboxB}{\usebox\myboxA}%
    \else
        \hskip -0.5\mylenA\rlap{\usebox\myboxA}{\hskip 0.5\mylenA\usebox\myboxB}%
    \fi}
\makeatother

\newcommand{\tens}[1]{\boldsymbol{\mathscr{#1}}}
\newcommand{\vect}[1]{\ensuremath{\mathbf{#1}}}
\newcommand{\mat}[1]{\ensuremath{\mathbf{#1}}}

\newcommand{\argmin}{\mathop{\rm argmin}}

\newcommand{\R}{\mathbb{R}}

\newcommand{\tA}{\tens{A}}

\newcommand{\tX}{\tens{X}}

\newcommand{\tE}{\tens{E}}

\newcommand{\tV}{\tens{V}}

\newcommand{\tG}{\tens{G}}

\newcommand{\tN}{\tens{N}}

\newcommand{\tY}{\tens{Y}}

\newcommand{\la}{\langle}
\newcommand{\ra}{\rangle}

\newcommand{\A}{\mat{A}}

\newcommand{\M}{\mat{M}}

\newcommand{\X}{\mat{X}}
\newcommand{\Y}{\mat{Y}}

\newcommand{\G}{\mat{G}}

\renewcommand{\H}{\mat{H}}

\newcommand{\mA}{\mat{A}}

\newcommand{\hmA}{\widehat{\mA}}

\newcommand{\x}{\vect{x}}

\newcommand{\partitle}[1]{\smallskip \noindent \textbf{#1.}}

\newcommand{\cmark}{\ding{51}}%
\newcommand{\xmark}{\ding{55}}%

\def\BibTeX{{\rm B\kern-.05em{\sc i\kern-.025em b}\kern-.08em
    T\kern-.1667em\lower.7ex\hbox{E}\kern-.125emX}}
\begin{document}

\title{Communication Efficient Generalized Tensor Factorization for Decentralized Healthcare Networks}

\author{\IEEEauthorblockN{Jing Ma$^1$, Qiuchen Zhang$^1$, Jian Lou$^{1,2}$, Li Xiong$^1$, Sivasubramanium Bhavani$^1$, Joyce C. Ho$^1$}
\IEEEauthorblockA{$^1$\textit{Emory University, Atlanta, Georgia}\\
$^2$\textit{Xidian University, Guangzhou}\\
\{jing.ma, qiuchen.zhang, jian.lou, lxiong, sivasubramanium.bhavani, joyce.c.ho\}@emory.edu, \{jlou\}@xidian.edu.cn}
}

\maketitle

\begin{abstract}
Tensor factorization has been proved as an efficient unsupervised learning approach for health data analysis, especially for computational phenotyping, where the high-dimensional Electronic Health Records (EHRs) with patients history of medical procedures, medications, diagnosis, lab tests, etc., are converted to meaningful and interpretable medical concepts. Federated tensor factorization distributes the tensor computation to multiple workers under the coordination of a central server, which enables jointly learning the phenotypes across multiple hospitals while preserving the privacy of the patient information. However, existing federated tensor factorization algorithms encounter the single-point-failure issue with the involvement of the central server, which is not only easily exposed to external attacks, but also limits the number of clients sharing information with the server under restricted uplink bandwidth. In this paper, we propose {\tt CiderTF}, a communication-efficient decentralized generalized tensor factorization, which reduces the uplink communication cost by leveraging a four-level communication reduction strategy designed for a generalized tensor factorization, which has the flexibility of modeling different tensor distribution with multiple kinds of loss functions.
Experiments on two real-world EHR datasets demonstrate that {\tt CiderTF} achieves comparable convergence with the communication reduction up to 99.99\%.
\end{abstract}


\maketitle

\IEEEdisplaynontitleabstractindextext

\IEEEpeerreviewmaketitle

\section{Introduction}\label{sec:introduction}
There has been growing interest in the use of Electronic Health Record (EHR), which typically creates digital records for patients indicating the details including diagnosis and treatment, prescription medications, lab tests, and other vital information which provides assistance to each patient.
The widespread adoption of EHR systems has facilitated the rapid accumulation of the patients' clinical data from numerous medical institutions -- enabling not only more accurate capturing of patients' medical information and clinical decision support \cite{ehrenstein2019obtaining}, but also improving workflow efficiency and patient care quality \cite{thakkar2006risks}. Yet, successfully mining the massive, high-dimensional EHR data is a challenging task due to sparse, missing, and noisy measurements \cite{miotto2016deep,weiskopf2013methods}. Computational phenotyping is the process of mapping the high-dimensional EHR data into meaningful medical concepts, which characterize a patient's clinical behavior and corresponding treatments, and can be further applied to multiple downstream tasks such as Genome-wide association studies (GWAS) \cite{liao2013associations,luo2017tensor} and risk prediction of a certain disease \cite{gastounioti2018incorporating}. Tensor factorization has been proven to be an efficient unsupervised learning approach to automatically extract phenotypes without the process of manual labeling \cite{ho2014marble,wang2015rubik, perros2017spartan}, and it is also widely applied to multidimensional data which is beyond the scenario of health data analysis such as spatio-temporal data analysis \cite{ma2020spatio}, network embedding \cite{ma2021temporal}, and computer vision \cite{vasilescu2002multilinear}. 

Traditional tensor factorization involves the use of the CANDECOMP/PARAFAC or canonical polyadic (CP) tensor factorization \cite{carroll1970analysis,harshman1970foundations} and its generalization \cite{hong2018generalized}, which are fundamental tools for analyzing multidimensional data in various domains. However, tensor factorization can only operate on data that can fit in the main memory, and the restricted scalability of tensor factorization has limited its application among large datasets.
This motivated the development of a variety of distributed tensor factorization algorithms \cite{kang2012gigatensor,beutel2014flexifact,choi2014dfacto}, where the large tensor is partitioned into multiple small tensors and thus parallelizes the tensor computation. Among these, federated tensor factorization \cite{kim2017federated,ma2019privacy,ma2021communication} has been developed as a special distributed tensor factorization paradigm. 
Beyond sharing similar computation and storage complexity, federated tensor factorization is able to preserve the data privacy by distributing the horizontally partitioned tensors to multiple medical institutions to avoid direct data sharing, and aims to learn the shared phenotypes through joint tensor factorization without communicating the individual-level data. Moreover, with the participation of different data sources, federated tensor factorization also helps mitigate the bias of analyzing data from single source, and achieves better generalizability.

Under the federated learning settings, the central server is the most important computation resource as it is in charge of coordinating the clients (i.e., picking random clients to communicate at each iteration), aggregating the clients' intermediate results, and updating the global model. This shows that federated learning systems heavily rely on the central server. However, a single server might have several shortcomings: 1) limited connectivity and bandwidth, which restrict the server from collecting data from as many clients as possible; 2) vulnerability to malfunctions, which will cause inaccurate model updates, or even learning failures; and 3) exposure to external attacks and malicious adversaries, which will perform physical and adversarial attacks to models \cite{madry2017towards, zhang2020broadening, finlayson2019adversarial}. This will lead to sensitive information leakage (thus, under the federated settings, a server is usually assumed to be honest-but-curious \cite{kim2017federated, ma2019privacy}). Due to the above limitations, it is obvious that traditional federated tensor factorization suffers from the bottleneck of the central server regarding the communication latency and bandwidth, and is exposed to high risk of single-point-failure.
To avoid relying on the server as the only source of computation, decentralization has been proposed as a solution to this single-point-failure issue \cite{li2020blockchain,lian2017can}. Decentralized federated learning is designed without the participation of the central server, while each client will rely on its own computation resources and communicate only with its neighbors in a peer-to-peer manner. 
Besides the necessities of a decentralized communication topology, it is also worth noting that the network capacity between clients are usually much smaller than the datacenter in many real-world applications \cite{vulimiri2015global}. Therefore it is necessary that the clients communicate the model updates efficiently with limited cost. 

In this paper, we study the decentralized optimization of tensor factorization under the horizontal data partition setting, and propose {\tt CiderTF}, a Communication-effIcient DEcentralized geneRalized Tensor Factorization algorithm for collaborative analysis over a communication network.  
To enable more flexibility on choosing different loss functions under various scenarios, we extend the classic federated tensor factorization into a more generalized tensor factorization. 
To the best of our knowledge, this paper is the first one proposing a decentralized generalized tensor factorization, let alone considering the decentralized setting with communication efficiency. Our contributions are briefly summarized as follows.

First, we develop a decentralized tensor factorization framework which employs four levels of communication reduction strategies to the decentralized optimization of tensor factorization to reduce the communication cost over the communication network. At the element-level, we utilize gradient compression techniques \cite{karimireddy2019error,zheng2019communication,basu2019qsparse,agarwal2018cpsgd,rothchild2020fetchsgd} to reduce the number of bytes transmitted between clients by converting the partial gradient from the floating point representation to low-precision representation. At the block-level, we apply the randomized block
coordinate descent \cite{beck2013convergence, fu2020block, nesterov2012efficiency} for the factor updates, which only requires sampling one mode from all modes of a tensor for the update per round and communicating only one mode factor updates with the neighbors. 
At the round-level, we adopt a periodic communication strategy \cite{stich2018local, lin2018don, basu2019qsparse} to reduce the communication frequency by allowing each client to perform $\tau>1$ local update rounds before communicating with its neighbors. In addition, at the communication event-level, we apply an event-triggered communication strategy \cite{liu2017asynchronous, du2018distributed, singh2020sparq} to boost the communication reduction at the round level. 

Second, we further incorporate Nesterov's momentum into the local updates of {\tt CiderTF} and propose {\tt CiderTF\_m}, in order to achieve better generalization and faster convergence. 

Third, we conduct comprehensive experiments on both real-world and synthetic datasets to corroborate the theoretical communication reduction and the convergence of {\tt CiderTF}. Experiment results demonstrate that {\tt CiderTF} achieves comparable convergence performance with the communication reduction of 99.99\%. Furthermore, we conduct an extensive case study on MIMIC-III data with regard to the factorization quality from both quantitative and qualitative aspects. The resulting factor matrices highly resembles the factors extracted by the centralized tensor factorization baseline with significantly less communication cost. The {\tt CiderTF} extracted phenotypes are shown to be highly interpretable according to a clinician.

\section{Preliminaries and Background}
In this section, we summarize the frequently used definitions and notations, and introduce the background knowledge of tensor factorization, the related communication reduction techniques, and decentralized optimization.

\subsection{Notations and Operators}
For a $D$-th order tensor $\tX\in\mathbb{R}^{I_1\times...\times I_D}$, the tensor entry indexed by $(i_1,...,i_D)$ is denoted by the MATLAB representation $\tX(i_1,...,i_D)$. Let $\mathcal{I}$ denote the index set of all tensor entries, $|\mathcal{I}| = I_{\Pi} = \prod_{d=1}^{D}I_d$. The mode-$d$ unfolding (also called matricization) is denoted by $\X_{<d>}\in \mathbb{R}^{I_{d}\times I_{\Pi}/I_d}$.

\begin{definition}
\label{def.mttkrp}
(\textit{MTTKRP}). The MTTKRP operation stands for the \textit{matricized tensor times Khatri-Rao product}. Given a tensor $\tY \in \mathbb{R}^{I_1\times...\times I_D}$, its mode-$d$ matricization is $\Y_{<d>}$, $[\A_{(1)},...,\A_{(D)}]$ is the set of CP factor matrices. $\H_d \in \mathbb{R}^{ I_{\Pi}/I_d \times R}$ is defined as
\begin{equation*}
    \H_d = \A_{(D)} \odot...\odot\A_{(d+1)}\odot\A_{(d-1)}...\odot\A_{(1)},
\end{equation*}
where $\odot$ is the Khatri-Rao product. The MTTKRP operation can thus be defined as the matrix product between $\Y_{<d>}$ and $\H_d$ as $\Y_{<d>}\cdot\H_{d}$.
\end{definition}

\begin{table}[]
\caption{\small Symbols and notations used in this paper}
\centering
\begin{small}
\begin{tabular}{c|c}
\hline
Symbol & Definition \\
\hline
 $\x,\X,\tX$& Vector, Matrix, Tensor\\
 $\tX_{<d>}$ & Mode-$d$ matricization of $\tX$\\
 $\|\cdot\|_1$ & $\ell _1$-norm\\
 $\|\cdot\|_F$ & Frobenius norm \\
  $\circledast$ & Hadamard (element-wise) multiplication\\
  $\odot$ & Khatri Rao product \\
  $\circ$ & Outer product \\
  $\la \cdot,\cdot \ra$ & Inner product \\
\hline
\end{tabular}
\end{small}
 \label{table.notations}
\vspace{-1em}
\end{table}

\subsection{CP Tensor Factorization and its Generalization}
\begin{definition}
(\textit{CANDECOMP-PARAFAC Decomposition}). 
The CANDECOMP-PARAFAC (CP) decomposition is to approximate the original tensor $\tX$ by the sum of $R$ rank-one tensors
\begin{equation}
	\tX \approx \tA = \sum _{i=1}^R \A_{(1)}(:,i)\circ...\circ\A_{(D)}(:,i),
\end{equation}
where $R$ is the rank of tensor $\tX$, $\A_{(d)}$ is the $d$-th mode factor matrix, $\A_{(d)}(:,i)$ is the $i$-th column of $\A_{(d)}$ representing the $i$-th latent component.
\end{definition}

Generalized CP (GCP) \cite{hong2018generalized} extends the classic CP by using the element-wise loss function to support other loss functions. The objective function of GCP is represented as
\begin{equation}
\begin{split}
        \arg\min _{\tA} & F(\tA,\tX)  = \sum_{i\in \mathcal{I}}f(\tA(i),\tX(i)) \\& ~s.t.~ \tA = \sum _{i=1}^R \A_{(1)}(:,i)\circ...\circ\A_{(D)}(:,i),
\end{split} 
\end{equation}
GCP not only preserves the low-rank constraints as CP decomposition, it also enjoys the flexibility of choosing different loss functions according to different data distributions by leveraging the elementwise objective function. For example, for data indexed by $i\in \mathcal{I}$ with Gaussian distribution, we use least square loss to model it, which in turn yields the classic CP decomposition:
\begin{equation}
\label{eq.square}
    f_{\text{square}}(\tA(i),\tX(i)) = (\tA(i)-\tX(i))^2.
\end{equation}
On the other hand, for binary data indexed by $i\in \mathcal{I}$, we can use Bernoulli-logit loss to fit it:
\begin{equation}
\label{eq.logit}
    f_{\text{logit}}(\tA(i),\tX(i)) = \log(1+\tA(i)) - \tX(i)\tA(i),
\end{equation}

\subsection{Communication Reduction}
\partitle{Gradient compression}
Communication can be a primary bottleneck of the efficiency of the distributed training, especially in federated learning since the connection between clients and the server usually operates at low speeds ($\sim$1 Mbps) (\cite{rothchild2020fetchsgd}), and the uplink bandwidth is generally slower than the downlink bandwidth. Gradient compression based methods can compress the communicated information that are transmitted from clients and the server by reducing the number of bits. 

\partitle{Periodic communication}
Periodic communication, which is also known as local SGD \cite{stich2018local, lin2018don}, has been developed in order to overcome the communication bottleneck in distributed training. Instead of keeping different clients in frequent synchronization, periodic communication allows clients to perform $\tau>1$ local updates before communicating, which reduces the communication frequency. Most recently, \cite{basu2019qsparse} explored the combination of periodic communication with various gradient compression strategies, \cite{stich2019error} introduced the error-feedback to analyze the convergence rate for the biased gradient compression and local SGD.

\partitle{Event-triggered communication}
The event-triggering mechanism was first proposed in the control community \cite{dimarogonas2011distributed, heemels2012introduction, seyboth2013event}, and then was extended to be used in distributed optimization \cite{chen2016event, du2018distributed}. Most recently, \cite{singh2020sparq, singh2020squarm} were proposed which study the combination of the event-driven lazy communication with gradient compression under the decentralized settings.

\subsection{Decentralized Optimization}
Decentralized optimization algorithms have been widely studied from multiple domains considering the limited bandwidth, communication latency, and data privacy of the distributed networks. 

In machine learning domain, D-PSGD \cite{lian2017can} was first proposed for the non-convex setting as a decentralized version of SGD with linear speedup, where $K$ is the number of workers, and $T$ is the total number of iterations. \cite{tang2018d} extended D-PSGD to the data heterogeneity settings. \cite{koloskova2020unified} provided a unified framework for the gossip-based decentralized SGD with theoretical convergence analysis for 

In addition, there are also multiple works investigating the communication reduction in decentralized optimization. \cite{tang2018communication} proposed DCD/ECD-PSGD which quantize the model updates with high precision quantizers. \cite{koloskova2019decentralized} proposed CHOCO-SGD, which is the first work that supports arbitrary compressors. \cite{lu2020moniqua} expanded the scope of the applicable quantizers and supported 1-bit quantizer with no additional memory required. \cite{singh2020sparq} further reduced the communication cost with an event-driven communication strategy.

\begin{figure} 
\centering
\includegraphics[width=0.4\textwidth]{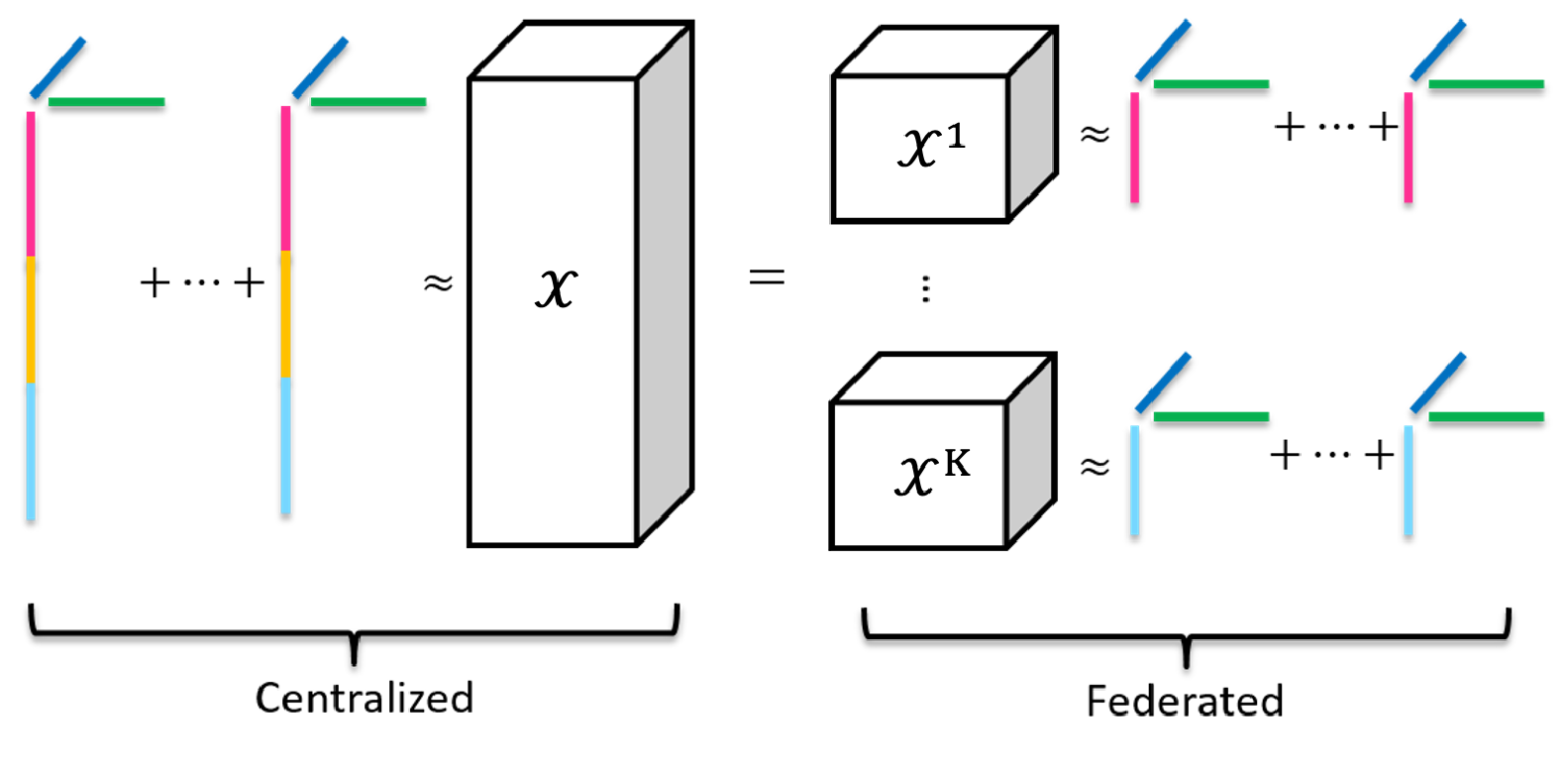}
\caption{Illustration of collaborative phenotyping via federated tensor factorization \cite{kim2017federated}.}
\label{fig:centralized_figure}
\end{figure}

\begin{figure}[htbp]
\centering
\includegraphics[width=0.45\textwidth]{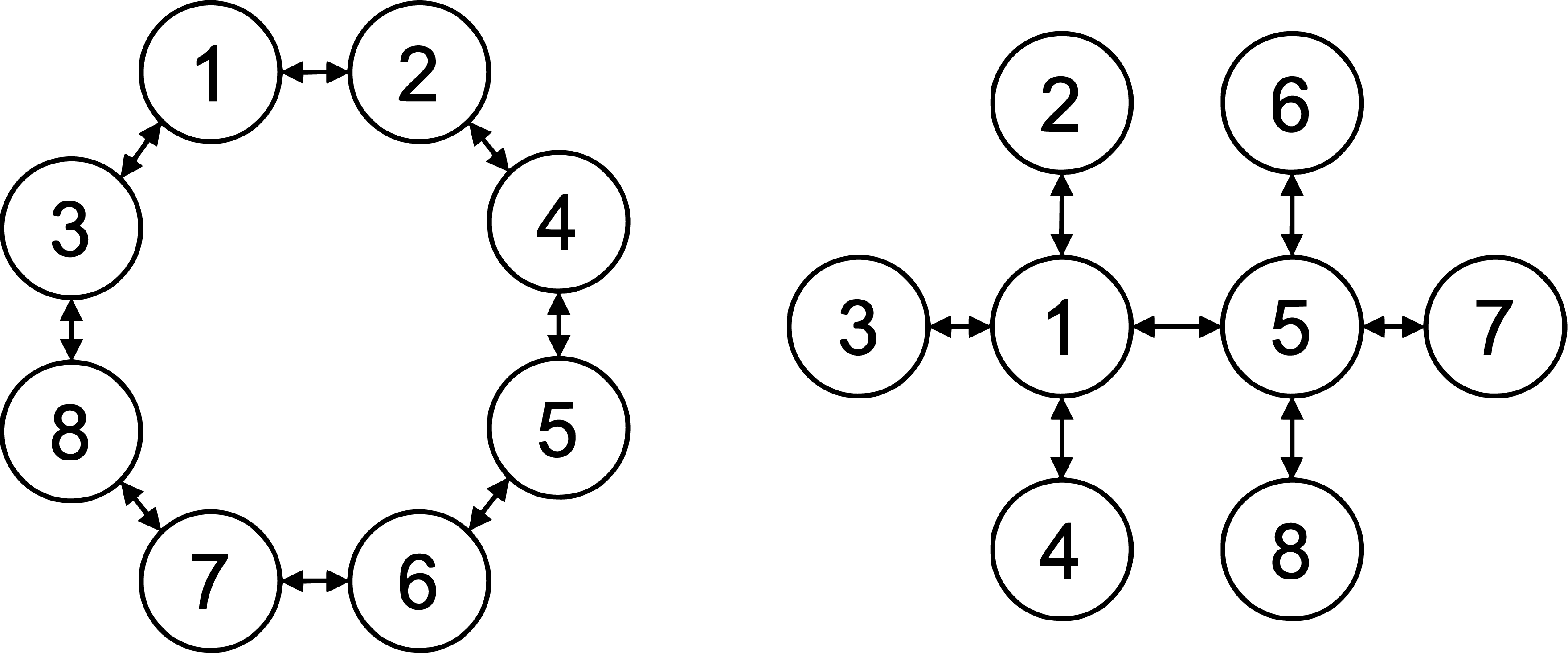}
\caption{Ring topology (left) and star topology (right).}
\label{fig:topology}
\end{figure}

\section{Proposed Method}
We study the decentralized generalized tensor factorization, where the EHR tensor $\tX\in\mathbb{R}^{I_1\times,...,\times I_D}$ is horizontally partitioned along the patient mode into $K$ small local tensors $\tX^k\in\R^{I_{1^k}\times I_2\times...\times I_D}, k=1,\cdots,K$, which are distributed among $K$ institutions. The aim is to collaboratively learn the phenotypes through communicating with the neighboring clients in the decentralized network without the coordination of the central server. We propose to solve the decentralized optimization using the gossip-based algorithm with multiple communication reduction techniques, including block randomization, gradient compression, periodic communication, and event-triggered communication.

\subsection{Problem Formulation}
In the decentralized tensor factorization setting, the communication topology is represented by an undirected graph $\tG=(\tV, \tE)$, where $\tV:=\{1, 2, ..., K\}$ denotes the set of clients participating in the communication network. Each node $k$ in the graph represents a client. The neighbors of client $k$ is denoted as $\tN_k:=\{(k,j):(k,j)\in \tE\}$. There is a connectivity matrix $W\in \mathbb{R}^{K\times K}$, the $(k,j)$-th entry $w_{kj}\in [0,1], \forall (k,j)\in \tE$ in which denotes the weights of edge $(k,j)\in \tE$ and measures how much the client $k$ is impacted by client $j$. If the graph is symmetric, then $w_{kj}=w_{jk}$, while if there is no connection between client $k$ and client $j$, then $w_{kj}=0$. We assume the connectivity matrix $W$ is symmetric and doubly stochastic where each row and column sums to one ($\sum_k w_{kj}=\sum_j w_{jk}=1$). 

Each client in the decentralized communication graph will hold a local tensor $\tX^k$, which can be seen as the horizontal partition of a global tensor $\tX$. The aim for the decentralized federated learning is to jointly factorize the local tensors $\tX^k$ to get the globally shared feature factor matrices $\A_{(2)},...,\A_{(D)}$, and the individual mode factor matrices $\A^k_{(1)}$ from all clients
\begin{equation}
\tX=\left[\begin{array}{c}
\tX^1\\
\vdots\\
\tX^K
\end{array}\right] \approx \left[\begin{array}{c}
\tA^1 = \A_{(1)}^1\circ\A_{(2)}\circ...\circ\A_{(D)}\\
\vdots\\
\tA^K =\A_{(1)}^K\circ\A_{(2)}\circ...\circ\A_{(D)}\\
\end{array}\right].
\end{equation}
The objective function for the decentralized generalized tensor factorization is shown as
 \begin{equation}
   \begin{split}
     & \argmin _{(\A_{(1)},...,\A_{(D)})} \sum _{k=1}^K F(\tA,\tX^k),
    \\
    & s.t.~\tA = \A_{(1)}\circ...\circ\A_{(D)},
  \end{split}
\end{equation}
which can be further extended to other multiblock optimization problems which are not limited to tensor factorization \cite{zeng2019global}.

\begin{algorithm}[tbp]
\small
\caption{{\tt CiderTF}: Decentralized Generalized Tensor Factorization with Compressed, Block-randomized, Periodic, and Event-triggered Communication}
\label{alg.decentralized.gtf.sgd}
\begin{algorithmic}[1] 
\REQUIRE Input tensor $\tX$, constant learning rate $\gamma[t]$, $\A[0], \A^k[0] = \A[0], \forall k=1,...,K$, randomized block sampling sequence $d_{\xi}[0],...,d_{\xi}[T]$, event-triggering threshold $\lambda[t]$;
\FOR{$t = 0,...,T$}
\STATE \textbf{On Each Client Nodes $k\in 1,...,K$:}
\IF{$d=d_{(\xi)}[t]$}
\STATE Compute stochastic gradient $\G_{(d)}^k[t]$ by eq. (\ref{eq.general.stochastic.grad.local});
\STATE $\A_{(d)}^k[t+\frac{1}{2}] = \A_{(d)}^k[t] - \gamma[t]\G_{(d)}^k[t]$; 
\IF{$(t \mod \tau) \neq 0$} 
\STATE No communication: \\
$\A_{(d)}^k[t+1] = \A_{(d)}^k[t+\frac{1}{2}]$, $\hmA_{(d)}^k[t+1] = \hmA_{(d)}^k[t]$; 
\ELSE
\FOR{$j\in\tens{N}_k \cup k$}
\IF{$\|\A_{(d)}^k[t+\frac{1}{2}] - \hmA_{(d)}^k[t]\|_F^2 \geq \lambda[t] (\gamma[t])^2$}
\STATE $\bm{\Delta}^k_{(d)}[t]=${\tt Compress}$(\A_{(d)}^k[t+\frac{1}{2}]-\hmA_{(d)}^k[t])$;
\ELSE
\STATE $\bm{\Delta}^k_{(d)}[t] = \bm{0}_{I^k\times R}$;
\ENDIF
\STATE Send $\bm{\Delta}^k_{(d)}[t]$ to all $j$ and receive $\bm{\Delta}^j_{(d)}[t]$ from all $j$, where $j\in \tens{N}_k$;
\STATE $\hmA_{(d)}^j[t+1] = \hmA_{(d)}^j[t] + \bm{\Delta}^j_{(d)}[t]$;
\ENDFOR
\STATE $\A_{(d)}^k[t+1] = \A_{(d)}^k[t+\frac{1}{2}] + \varrho \sum_{j\in\tens{N}^k} w_{kj} (\hmA_{(d)}^j[t+1] - \hmA_{(d)}^k[t+1])$;
\ENDIF
\ELSIF{$d\neq d_{\xi}[t]$} 
\STATE $\A_{(d)}^k[t+1] = \A_{(d)}^k[t]$, $\hmA_{(d)}^k[t+1] = \hmA_{(d)}^k[t]$;
\ENDIF
\ENDFOR
\end{algorithmic}
\end{algorithm}

\subsection{{\tt CiderTF}: Decentralized Generalized Tensor Factorization with Compressed, Block-randomized, Periodic, and Event-triggered Communication}

\subsubsection{Overview}
We propose {\tt CiderTF}, a decentralized tensor factorization framework which achieves communication efficiency through four levels of communication reduction. We utilize the widely used sign compressor \cite{stich2019error, stich2018sparsified} for gradient compression. 

\begin{definition}
\label{def.sign.compressor}
(Sign Compressor) For an input tensor $\x\in\mathbb{R}^d$, its compression via ${\tt Sign}(\cdot)$ is
    ${\tt Sign}(\x) = \|\x\|_1/d \cdot sign(\x)$, where $sign$ takes the sign of each element of $\x$.
\end{definition} 

Moreover, we apply the block randomization for the factor updates, which can not only reduce the computational complexity per iteration, but also eliminate the need of sending full factor matrices. 
Furthermore, we reduce the communication frequency via the combination of the periodic communication and the event-triggered communication strategies. Each client only need to check for the triggering condition every $\tau$ iterations, and will only need to communicate the compressed updates when the triggering condition is satisfied.

The detailed algorithm is shown in Algorithm \ref{alg.decentralized.gtf.sgd} with the key steps annotated. In {\tt CiderTF}, each client $k\in [K]$ maintains the local factor matrices $\A_{(d)}^k$ from each mode $d=1,...,D$. The goal is to achieve consensus on the feature mode factor matrices $\A_{(d)}^k, \forall d=2,...,D$. Therefore, besides the local factor matrices, each client also need to maintain the estimation of the local factor matrices $\A_{(d)}^j$ from both itself $k$ and its neighbors $\tens{N}_k$ ($j\in\tens{N}_k \cup k$). The sequence of the randomized sampling blocks for every round $t=1,...,T$ is denoted as $d_{\xi}[0],...,d_{\xi}[T]$. At every round for the sampled block $d_{\xi}[t]$, each client checks for the triggering condition for every $\tau$ iterations at the communication round (line 10). The triggering threshold is set to be $\lambda[t]$. When the difference between the updated factor and the local estimation is larger than the threshold, each client will send and receive the compressed updates to its neighbors. While if the triggering condition is not satisfied, then the clients will just communicate a matrix of zero instead (line 10-14). After receiving the compressed updates from all its neighbors, each client will first update the local estimation of the factor matrices $\hmA_{(d)}^j[t+1], j\in\tens{N}_k \cup k$ (line 16), and conduct the consensus step and update the local factors $\A_{(d)}^k[t+1]$ through the decentralized consensus step (line 18). At the non-communication round, each client will just keep updating the local factor matrices (line 6-7). For the rest of the blocks not selected, they will remain the same at the last round (line 20-22). 

\subsubsection{Optimization}
At each iteration, each client $k$ first need to compute the GCP gradient as the partial derivative with regard to the factor matrix $\A_{(d)}^k$ using the MTTKRP operator
\begin{equation}
    {\frac{\partial F(\tA^k,\tX^k)}{\partial \A_{(d)}^k[t]}} = \Y_{<d>}^k\H_d^k,
\end{equation}
where each element of the matrix $\Y_{<d>}$ is defined as
\begin{equation}
    \tY^k(i) = \frac{\partial f(\tA^k(i),\tX^k(i))}{\partial \tA^k(i)}, \forall i\in \mathcal{I},
\end{equation}
and $\H_d^k$ denotes the Khatri-Rao product of mode $d$ of the factor matrices as is shown in definition \ref{def.mttkrp}. Then the local factor matrix is updated with the gradient descent step
\begin{equation}
    \A_{(d)}^k[t+\frac{1}{2}] = \A_{(d)}^k[t] - \gamma[t]{\frac{\partial F(\tA^k,\tX^k)}{\partial \A_{(d)}^k[t]}}.
\end{equation}

\partitle{Fiber Sampling}
Note that the computational complexity of the full gradient ${\frac{\partial F(\tA^k,\tX^k)}{\partial \A_{(d)}^k[t]}}$ is $O(R \prod_{d=1}^D I_d)$, which is the bottleneck of the traditional gradient based optimization for tensor factorization, especially for EHR tensors where each dimension can be very large. 
In order to tackle the time complexity of computing the gradient, we propose to utilize an efficient fiber sampling technique \cite{battaglino2018practical, fu2020block}, which randomly samples $|\mathcal{S}_d|$ fibers from mode $d$.
The fiber sampling technique provides efficient computation of $\H_d^k(\mathcal{S}_d,:)$, where only fibers are sampled, and $\H_d^k(\mathcal{S}_d,:)$ is computed without forming $\H_d^k$ explicitly, but only the $s$-th rows of $\H_d^k$ are required to be computed as the Hadamard product ($\circledast$) of the certain rows of the factor matrices at time $t$ as $\H_d^k(s,:) = \A^k_{(1)}(i_1^s,:) \circledast...\circledast\A^k_{(d-1)}(i_{d-1}^s,:) \circledast \A^k_{(d+1)}(i_{d+1}^s,:)\circledast...\circledast\A^k_{(D)}(i_D^s,:)$,
where the indices of rows of the factor matrices are obtained from the index mapping $\{i_1^s,...,i_D^s\}, s\in \mathcal{S}_d$. Fiber sampling also allows us to avoid forming the full matricization of $\Y_{<d>}$, but only need to form $\Y_{<d>}^k(:,\mathcal{S}_d)$ with size $I_d \times |\mathcal{S}|$ \cite{kolda2019stochastic}.
Therefore, the local partial stochastic gradient, which is denoted as $\G^k_{(d)}[t]$, is considered as an unbiased estimation of the gradient ${\frac{\partial F(\tA^k,\tX^k)}{\partial \A_{(d)}^k[t]}}$, and can be efficiently computed with the fiber sampling technique as
\begin{equation}
\label{eq.general.stochastic.grad.local}
    \G^k_{(d)}[t] = \Y^k_{<d>}(:,\mathcal{S}_d)\H^k_d(\mathcal{S}_d,:),
\end{equation}

\partitle{Block randomization}
Besides fiber sampling, we utilize the block randomization \cite{fu2020block} to further improve the computation efficiency. Under block randomization, we randomly select a mode to update at each round, instead of updating all modes. In other words, for every epoch, there is a random variable $d_{\xi}[t]\in \{1, ..., D\}$ representing the selected mode, and the probability of each mode to be updated at each round is 
\begin{equation}
    Pr(d_{\xi}[t]=d)={\frac{1}{D}}.
\end{equation}
Specially for {\tt CiderTF}, we always keep the patient mode (i.e. the 1-st mode) securely at local to avoid directly sharing patient related information, thus when $d_{\xi}[t] = 1$, we skip the communication of this round, and only update the local patient mode factors. This not only improves the computation efficiency of the optimization, but also reduces the communication cost at the block level.

\subsection{{\tt CiderTF\_m}: {\tt CiderTF} with Nesterov's momentum}
We further propose {\tt CiderTF\_m} with Nesterov's momentum incorporated in the local SGD update step to speedup the convergence and achieve less total communication bits. After computing the partial stochastic gradient $\G^k_{(d)}[t]$ (line 4), we update the momentum velocity component as
\begin{equation}
    \M_{(d)}^k[t] = \G_{(d)}^k[t] + \beta \frac{\eta [t-1]}{\eta [t]}\M_{(d)}^k[t-1]
\end{equation}
where $\beta$ is the momentum parameter. The intermediate factor matrix will be updated as 
\begin{equation}
    \A_{(d)}^k[t+\frac{1}{2}] = \A_{(d)}^k[t] - \gamma[t](\G_{(d)}^k[t]+\beta \M_{(d)}^k[t])
\end{equation}

\subsection{Complexity Analysis}
We analyze the complexity from the perspective of computation, communication, and memory cost. 
\subsubsection{Computational Complexity}
\label{sec:computation_complexity}
\begin{theorem}
  The per-iteration computational complexity of \textbf{{\tt CiderTF}} for each client is $O(\frac{1}{D}(\sum_{d=1}^D I_d)R|\mathcal{S}|)$.
\end{theorem}
For each client, assume $|\mathcal{S}_d|$ are the same for all $d$ as $|\mathcal{S}|$, the computational complexity per-iteration of the partial stochastic gradient $\G^k_{(d)}[t] =\Y^k_{<d>}(:,\mathcal{S}_d)\H^k_d(\mathcal{S}_d,:)$ consists of three parts: 1) computing the matricization with fiber sampling $\Y^k_{<d>}(:,\mathcal{S}_d)$ takes $O(I_{d}|\mathcal{S}|)$; 2) computing the Khatri-Rao product with fiber sampled factor matrices $\H^k_d(\mathcal{S}_d,:)$ takes $O(R|\mathcal{S}|(D-1))$; 3) the matrix multiplication between $\Y^k_{<d>}(:,\mathcal{S}_d)$ and $\H^k_d(\mathcal{S}_d,:)$ takes $O(|\mathcal{S}|R I_d)$. In addition, the factor matrix updates is conducted per iteration with the size of $O(I_d R)$, where $I_d$ is the number of element of mode $d$, and $R$ is the rank of the decomposed tensor. Overall, the total computational complexity is summarized as $O(\frac{1}{D}(\sum_{d=1}^D I_d)R|\mathcal{S}|)$.

\subsubsection{Communication Complexity}
\begin{theorem}
     \textbf{{\tt CiderTF}} reduces a lower bound of $1-\frac{1}{32D\tau}$ communication.
\end{theorem}
The use of ${\tt Sign}$ compressor will require each client $k$ sending $\frac{1}{D}R\sum _{d=2}^D I_d$ bits to each neighbor in $\tens{N}_k$ with the block randomization in reducing the communication by a factor of $D$. The periodic communication strategy helps reduce the communication cost by ${\frac{1}{\tau}}$ and results in a total cost of $\frac{1}{D\tau}R\sum _{d=2}^D I_d$. Then without the event-triggering mechanism, there is a lower bound of communication reduction of $1-\frac{1}{32D\tau}$, compared with the decentralized SGD with full precision gradients which has the per-iteration cost of $32(\sum_{d=1}^{D}I_d)$. The event-triggering mechanism helps further reduce the communication cost by an upper bound of $36\times$ per epoch (an epoch contains 500 iterations for our experiments). The total communication reduction is 99.99\% compared with the full precision decentralized SGD based on experimental results.

\subsubsection{Memory Complexity}
\begin{theorem}
     \textbf{{\tt CiderTF}} has the memory complexity of $O(|\mathcal{S}|\frac{1}{D}\sum_{d=1}^D I_d)$.
\end{theorem}
The memory complexity savings of {\tt CiderTF} comes from the fiber sampling technique, which eliminates the need for each client to form the whole mode-$d$ matricized tensor $\Y_{<d>}^k$ with the size of $\prod_{d=1}^D I_d$. Instead, each client only need to form a ``sketched version" of $\Y_{<d>}^k(:,\mathcal{S}_d)$ with size $I_d\times |\mathcal{S}|$. Thus the memory complexity is reduced to $O(|\mathcal{S}|\frac{1}{D}\sum_{d=1}^D I_d)$.

\section{Experiment}

\subsection{Experimental Settings}
\subsubsection{Datasets}
We conduct experiments on two real-world datasets, including MIMIC-III \cite{johnson2016mimic} and CMS \cite{cms}, which are large volume, publicly available and de-identified. 
MIMIC-III data contains more than fifty thousand of intensive care unit (ICU) stays from 2001-2012. The CMS (DE-SynPUF) dataset is a realistic set of claim data with the highest degree of protection on the patient information and has similar data structure as the real CMS data. It contains more than six billion beneficiary records from 2008-2010. We also generate a synthetic dataset with similar sparsity as the real-world datasets to further testify the generalizability of our algorithm. To reduce the sparsity, we follow the rules in \cite{kim2017federated} and select the top 500 diagnoses, procedures, and medications of the most frequently observed records to form the tensors with patient mode 34,272, 125,961, and 4000 for MIMIC-III, CMS, and Synthetic data, respectively.

\subsubsection{Baselines}
We consider the centralized tensor factorization baselines. i) \textbf{GCP} \cite{kolda2019stochastic} as the centralized baseline of the generalized tensor factorization; ii) \textbf{BrasCPD} \cite{fu2020block} as the computation efficient centralized tensor factorization baseline; iii) \textbf{Centralized {\tt CiderTF}}, {\tt CiderTF} with $K=1$ and uses error-feedback to adjust the compression error.

In addition, we also implement the decentralized version SGD under the non-convex settings as the decentralized baselines, since there is no existing decentralized tensor factorization framework. i) \textbf{D-PSGD} \cite{lian2017can, koloskova2020unified} as a pure decentralized version of stochastic gradient descent (SGD); ii) \textbf{SPARQ-SGD} \cite{singh2020sparq} as a decentralized communication-efficient stochastic gradient descent framework which employs the gradient compression, local SGD, and the event-triggered communication to reduce the communication cost; iii) \textbf{D-PSGDbras} can be considered as D-PSGD with block randomization.

\subsubsection{Parameter Settings}
Experiments are performed on two kinds of objective functions including Bernoulli-logit loss to fit the binary data (eq. \ref{eq.logit}) and Least Square Loss to fit the data with Gaussian distribution (eq. \ref{eq.square}), which is also considered as standard CP decomposition.

We set the number of iterations per epoch as 500. We use a fixed learning rate $\gamma[t]$, which is determined through searching the grid of powers of 2. We follow the rule in \cite{singh2020sparq} to set the triggering threshold $\lambda[t]$. The triggering threshold is initialized as $\lambda[0]$ at the beginning, and will be increased by a constant factor $\alpha_\lambda$ every $m$ epochs until convergence in order to prevent the clients satisfying the triggering condition every epoch. We set $\lambda[0]$ as ${\frac{1}{\gamma[t]}}$ according to \cite{singh2020sparq}, and set $\alpha_\lambda$ and $m$ through grid search within $[1,2]$ and $[1,5]$. For {\tt CiderTF\_m}, we set the momentum factor $\beta$ as 0.9.

\subsection{Result Analysis}
We form a decentralized communication topology as a ring, and have a default of eight workers with data horizontally partitioned and distributed evenly across all the eight clients. 

\subsubsection{Comparison to the Baselines}
\begin{figure*}[htbp]
\centering
\includegraphics[width=1\textwidth]{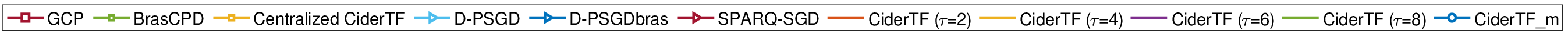}
\hspace{-0.75em}
\includegraphics[width=0.245\textwidth]{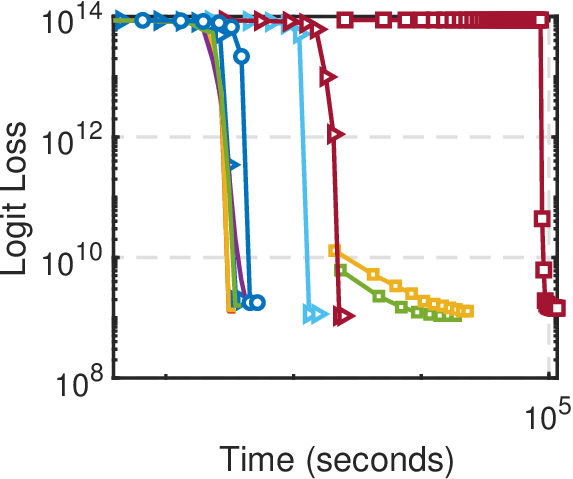}
\hspace{-0.75em}
\includegraphics[width=0.245\textwidth]{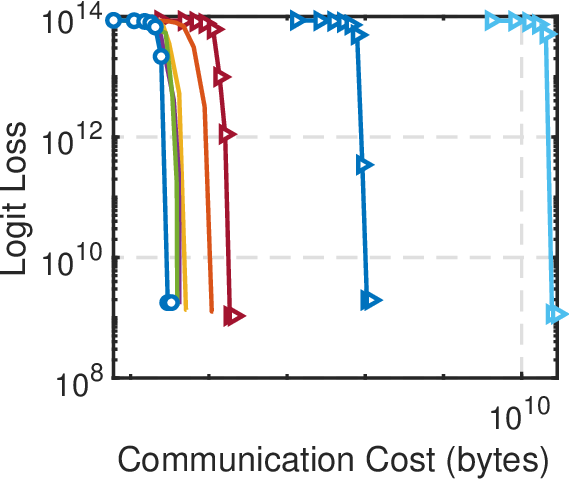}
\hspace{-0.75em}
\includegraphics[width=0.245\textwidth]{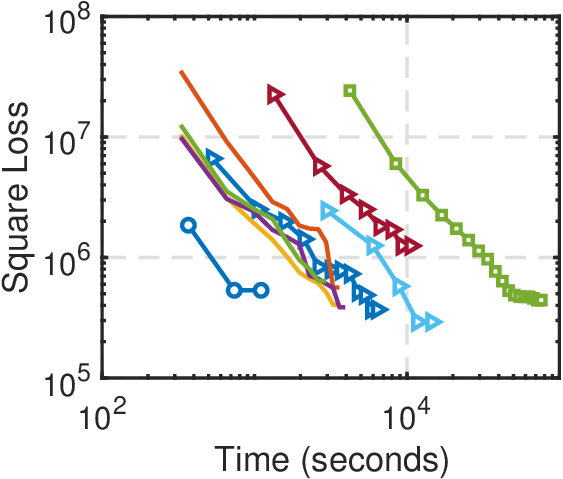}
\hspace{-0.75em}
\includegraphics[width=0.245\textwidth]{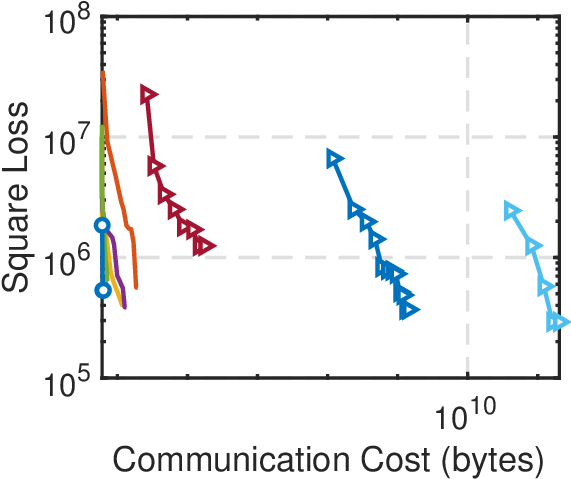}
\hspace{-0.75em}
\includegraphics[width=0.245\textwidth]{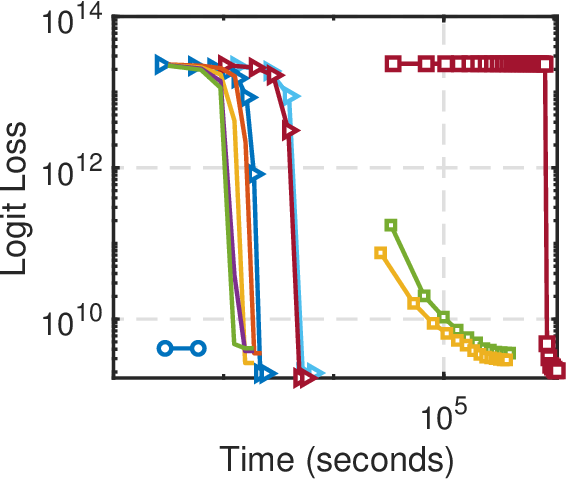}
\hspace{-0.75em}
\includegraphics[width=0.245\textwidth]{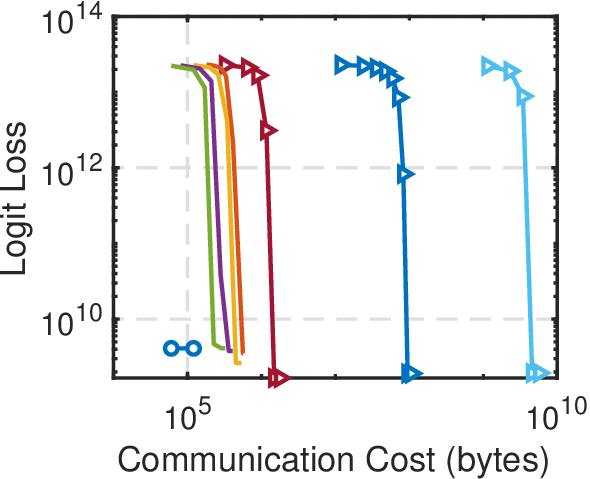}
\hspace{-0.75em}
\includegraphics[width=0.245\textwidth]{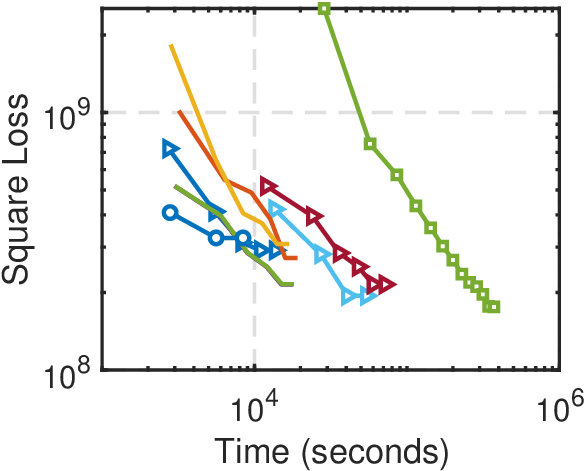}
\hspace{-0.75em}
\includegraphics[width=0.245\textwidth]{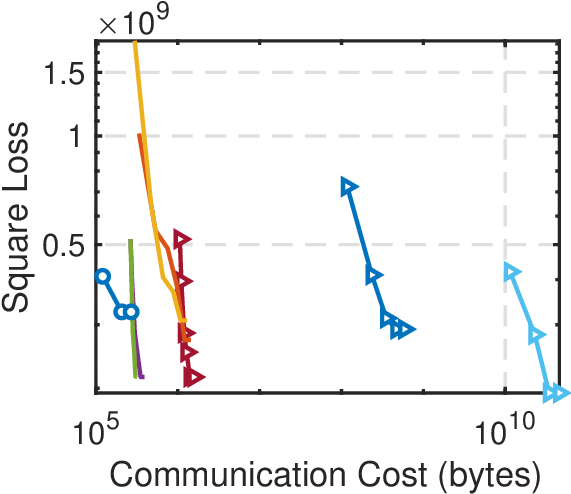}
\hspace{-0.75em}
\includegraphics[width=0.245\textwidth]{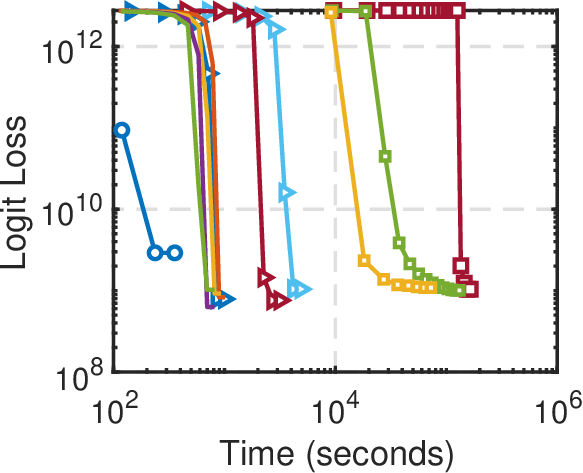}
\hspace{-0.75em}
\includegraphics[width=0.245\textwidth]{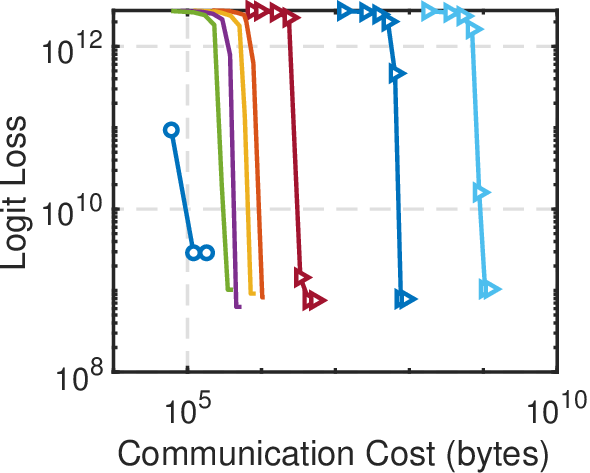}
\hspace{-0.75em}
\includegraphics[width=0.245\textwidth]{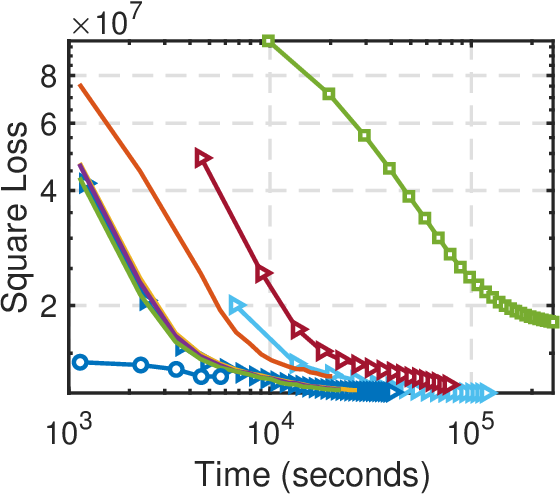}
\hspace{-0.75em}
\includegraphics[width=0.245\textwidth]{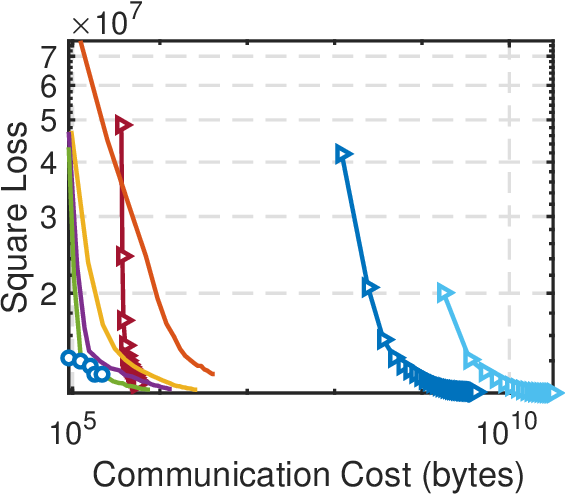}
\hspace{-0.75em}
\caption{Bernoulli-logit Loss (1-2 columns) and Least Square Loss (3-4 columns) for ring topology with respect to time and communication. Centralized approaches are marked with square, decentralized baselines are marked in triangle, {\tt CiderTF} with different number of local rounds are marked in solid lines. From top to bottom: CMS, MIMIC-III, Synthetic dataset.}
\label{fig:loss}
\end{figure*}

From fig. \ref{fig:loss}, we have four major observations. 

i) \textbf{{\tt CiderTF} converges to comparable losses as the centralized tensor factorization baselines.} {\tt CiderTF}, with various number of local update rounds ($\tau=\{2,4,6,8\}$), achieves similar losses at convergence compared with the centralized tensor factorization algorithms. These results empirically validate the convergence of {\tt CiderTF}.

ii) \textbf{{\tt CiderTF} has less communication cost compared with the decentralized baselines.} To achieve the same loss, {\tt CiderTF} takes 99.99\% less communication cost than D-PSGD, 75\% less communication cost than SPARQ-SGD and 99.92\% less than D-PSGDbras. This communication reduction is achieved without sacrificing the convergence rate compared with the decentralized SGD with full precision gradients. 

iii) \textbf{{\tt CiderTF} is computationally efficient.} From the 1, 3 columns of fig. \ref{fig:loss}, we observe that {\tt CiderTF} is computationally efficient compared with GCP and D-PSGD. This is because the fiber sampling technique and the block randomization helps reduce the computational complexity, which also verifies the computational complexity analysis in Sec. \ref{sec:computation_complexity}. {\tt CiderTF} is also slightly more efficient than BrasCPD mainly due to the scalability of the decentralized data distribution which helps parallelize the local tensor factorization.

iv) \textbf{Nesterov's momentum can offer {\tt CiderTF\_m} faster convergence, thus will lead to less overall communication cost}. From fig. \ref{fig:loss}, we observe that {\tt CiderTF\_m} requires less epochs to converge, which in turn helps reduce the total communication bytes with little sacrifice of the accuracy.

\subsubsection{Impact of Topology}
\begin{figure*}[htbp]
\centering
\includegraphics[width=0.98\textwidth]{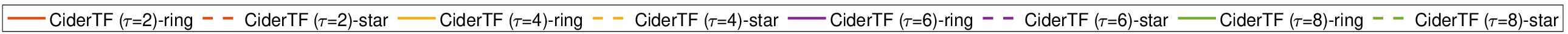}
\hspace{-0.75em}
\includegraphics[width=0.245\textwidth]{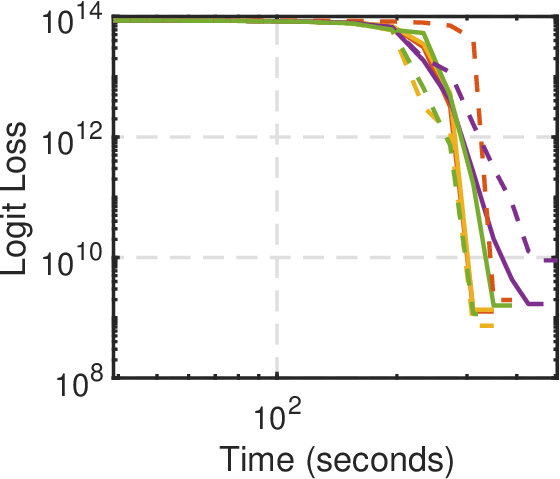}
\hspace{-0.75em}
\includegraphics[width=0.245\textwidth]{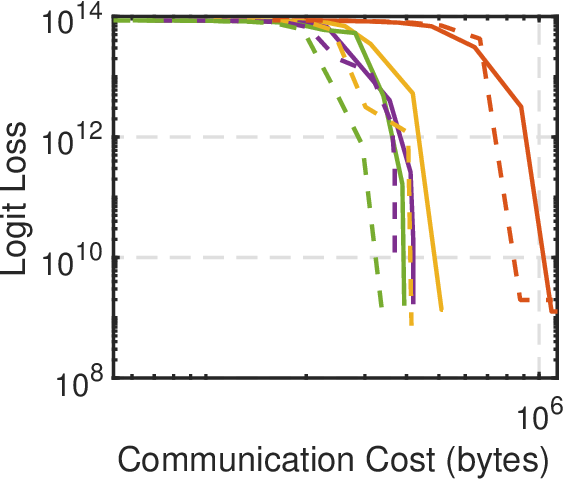}
\hspace{-0.75em}
\includegraphics[width=0.245\textwidth]{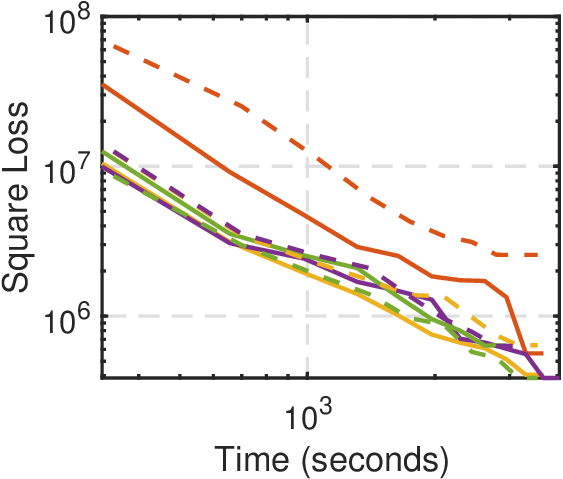}
\hspace{-0.75em}
\includegraphics[width=0.245\textwidth]{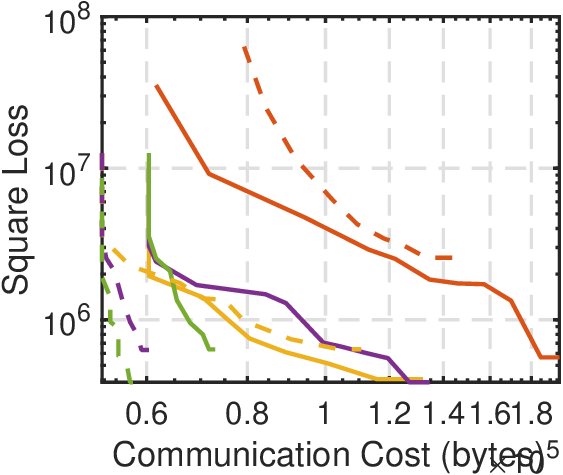}
\hspace{-0.75em}
\includegraphics[width=0.245\textwidth]{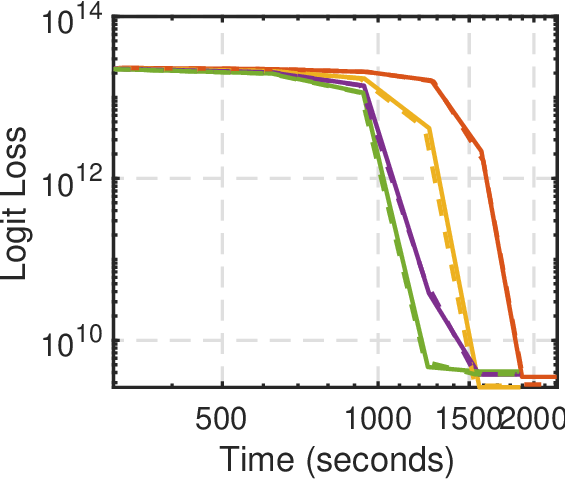}
\hspace{-0.75em}
\includegraphics[width=0.245\textwidth]{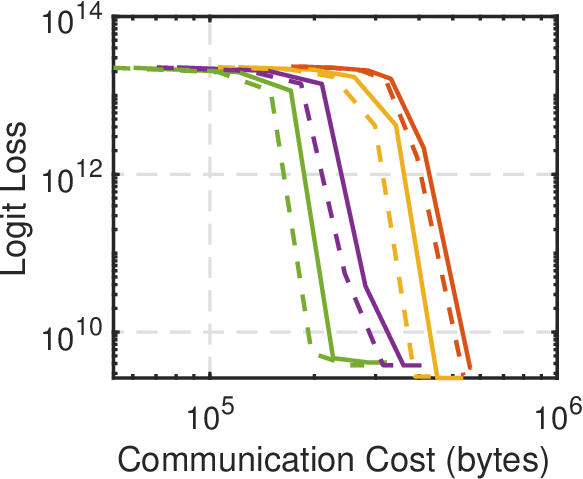}
\hspace{-0.75em}
\includegraphics[width=0.245\textwidth]{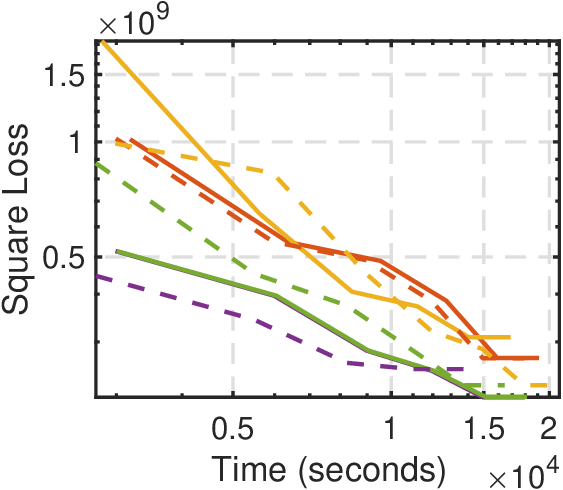}
\hspace{-0.75em}
\includegraphics[width=0.245\textwidth]{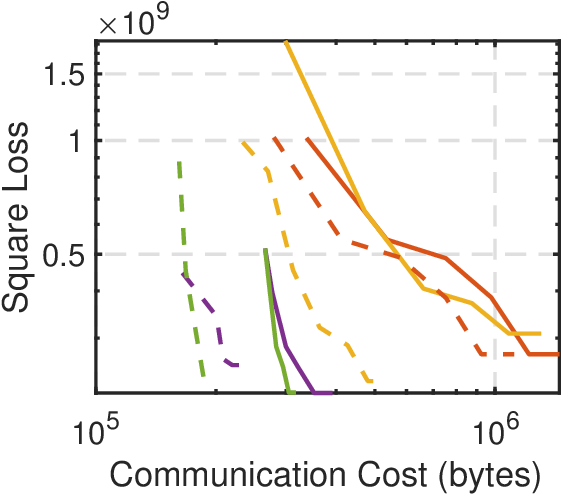}
\hspace{-0.75em}
\includegraphics[width=0.245\textwidth]{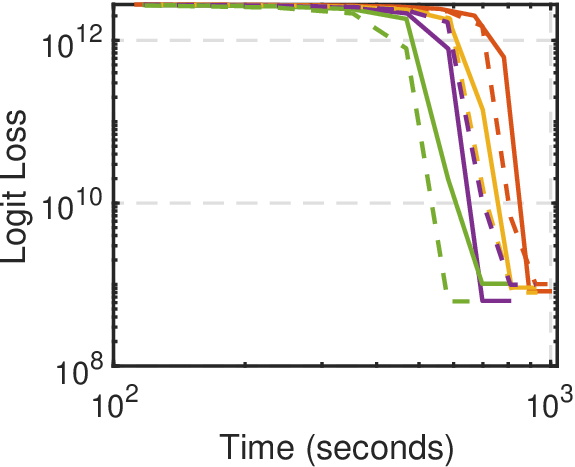}
\hspace{-0.75em}
\includegraphics[width=0.245\textwidth]{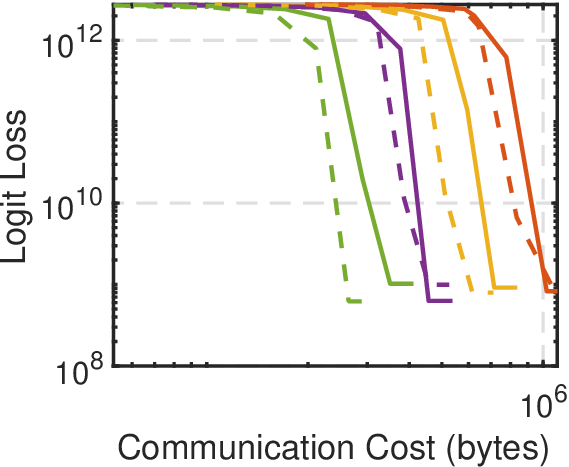}
\hspace{-0.75em}
\includegraphics[width=0.245\textwidth]{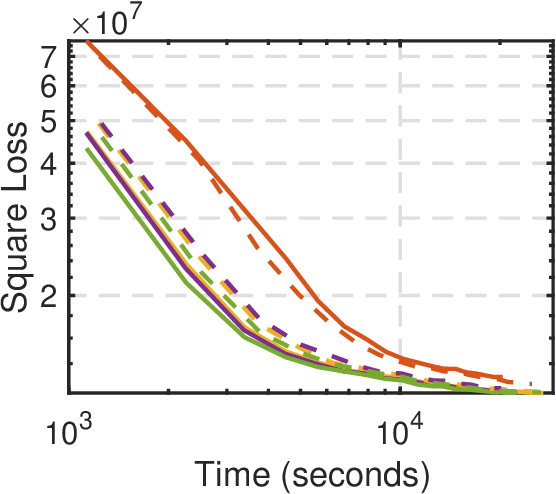}
\hspace{-0.75em}
\includegraphics[width=0.245\textwidth]{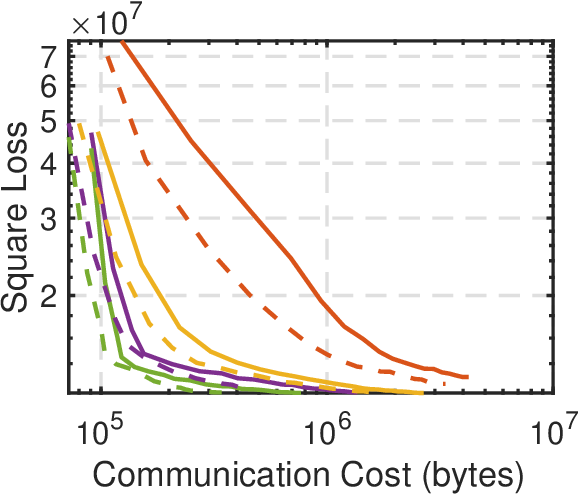}
\hspace{-0.75em}
\caption{Bernoulli-logit Loss (1-2 columns) and Least Square Loss (3-4 columns) for ring topology (solid lines) and star topology (dashed lines) with respect to time and communication. From top to bottom: CMS, MIMIC-III, Synthetic dataset.}
\label{fig:topology_loss}
\end{figure*}

We also test {\tt CiderTF} on two different topologies, including ring topology and the star topology for the same number of workers (shown in fig. \ref{fig:topology}). From the results in fig. \ref{fig:topology_loss}, we observe that different topologies do not affect the convergence performance and both of them converge to similar losses for both Bernoulli-logit Loss and Least Square Loss, which means that {\tt CiderTF} can generalize to different kinds of communication network topologies. Fig. \ref{fig:topology_loss} (left) illustrates that two topologies do not vary much in the computation time, since the number of workers are the same. However, star topology enjoys less communication cost because the total degree of the star topology is less than the ring topology (fig. \ref{fig:topology_loss} right).

\subsubsection{Scalability}
\begin{figure}[htbp]
\centering
\includegraphics[width=0.48\textwidth]{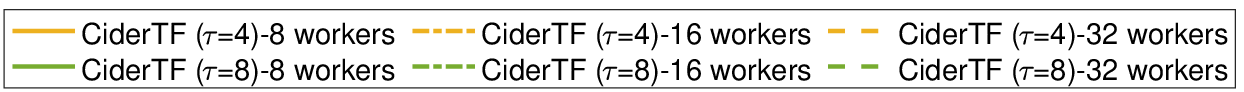}
\includegraphics[width=0.24\textwidth]{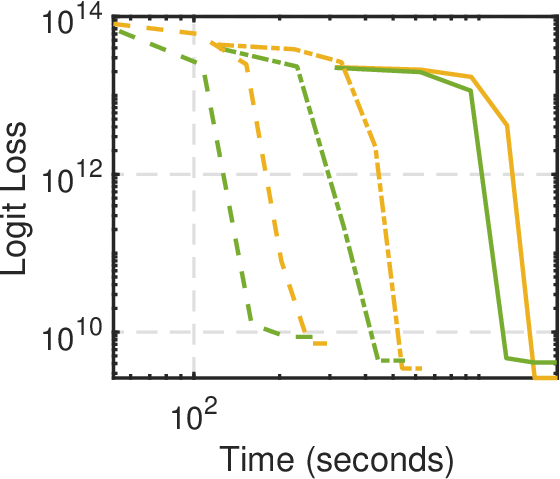}
\hspace{-0.75em}
\includegraphics[width=0.245\textwidth]{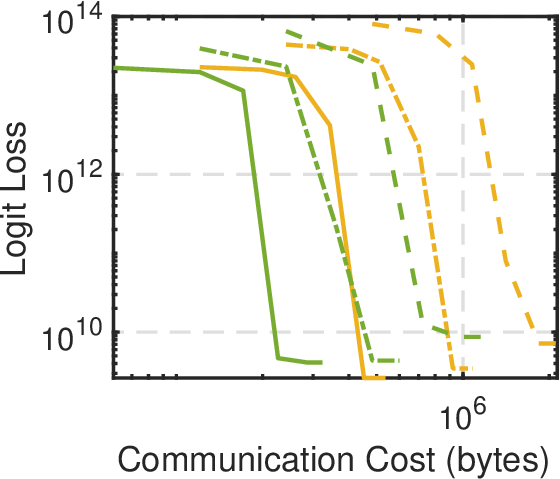}
\caption{Bernoulli-logit loss with respect to time and communication for MIMIC-III data with 8, 16, and 32 workers for local update rounds $\tau=4,8$.}
\label{fig:multiworker}
\end{figure}

Moreover, we test the scalability of {\tt CiderTF}. By increasing the number of clients from $K=8$ to $K=16$ and $K=32$ involved in the computation, we can observe linear scalability in the computation time (fig. \ref{fig:multiworker} left) without sacrificing the accuracy. However, as the number of clients increases, the communication cost will increase accordingly (fig. \ref{fig:multiworker} right). Therefore, there exists a computation-communication trade-off when increasing the number of clients involved in the decentralized tensor factorization framework.

\subsubsection{Ablation Study}
\begin{table*}[htbp]
\small
\caption{Comparison of the decentralized optimization algorithms in ablation study.}
  \setlength{\tabcolsep}{2pt}
  \centering
  \begin{tabular}{l c c c c c}
    \toprule
    \textbf{Algorithm} & \textbf{Element-level}  & \textbf{Block-level}& \textbf{Round-level} & \textbf{Event-driven} & \textbf{Compression Ratio}\\
    \midrule
    D-PSGD & \xmark & \xmark & \xmark & \xmark & 0 \\
    D-PSGDbras & \xmark & \cmark & \xmark & \xmark  & $1-1/D$\\
    D-PSGD+signSGD & \cmark & \xmark & \xmark  & \xmark & $1-1/32$ \\
    D-PSGDbras+signSGD & \cmark & \cmark & \xmark  & \xmark & $1-1/32 D$\\
    SPARQ-SGD & \cmark & \xmark & \cmark  & \cmark &  $1-1/32\tau$ \\
    {\tt CiderTF} & \cmark & \cmark & \cmark  & \cmark& $1-1/32D\tau$\\
    \bottomrule
\end{tabular}
\label{tab:ablation}
\end{table*}

We conduct an ablation study to clarify the reduction of the communication cost. We compare {\tt CiderTF} with the following algorithms: 1) D-PSGD: decentralized SGD with full gradient and full block and single round communication; 2) D-PSGDbras: D-PSGD with block randomization; 3) D-PSGD+signSGD: D-PSGD with gradient compression using sign compressor; 4) D-PSGDbras+signSGD: D-PSGD with block randomization and gradient compression. The algorithm comparisons are summarized in table \ref{tab:ablation}.

\begin{figure}[htbp]
\centering
\includegraphics[width=0.48\textwidth]{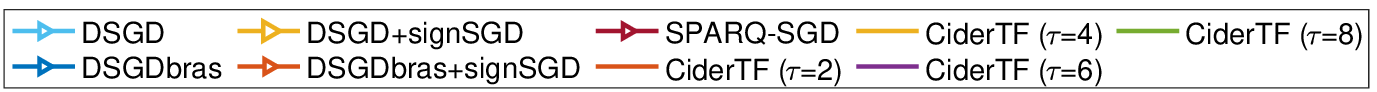}
\includegraphics[width=0.245\textwidth]{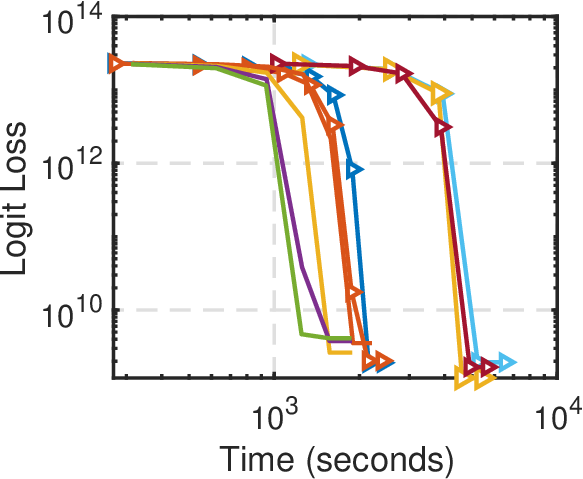}
\hspace{-0.75em}
\includegraphics[width=0.245\textwidth]{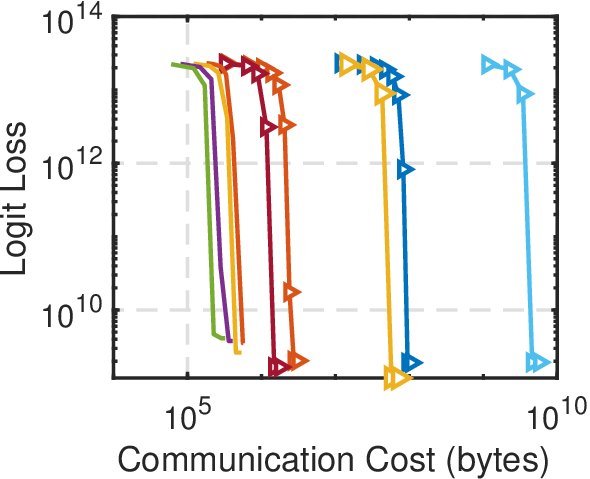}
\caption{Ablation study for decentralized optimization algorithms with different levels of communication reduction.}
\label{fig:ablation}
\end{figure}

From fig. \ref{fig:ablation}, we observe that D-PSGD which communicates the full gradient and blocks has the highest communication cost. Gradient compression plays the most important role in reducing the communication cost, which reduces the actual communication cost of MIMIC-III data of 96.88\%. The block randomization further reduces the communication cost to 75.00\%. The application of the event-driven and periodic communication helps reduce the communication with a lower bound of 87.5\% and up to an upper bound to 97.22\%.

\subsection{Case Study on MIMIC-III}

\begin{figure}
\centering
\includegraphics[width=0.48\textwidth]{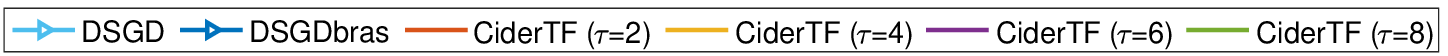}
\includegraphics[width=0.245\textwidth]{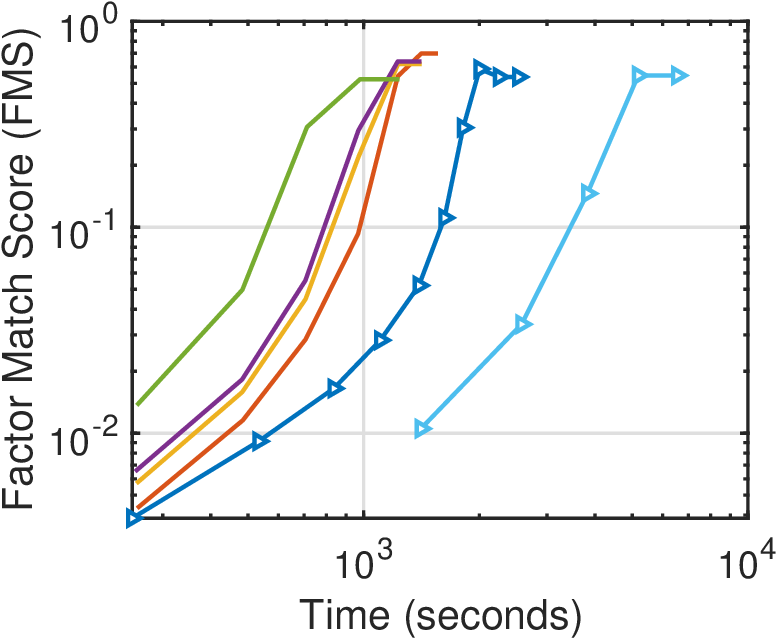}
\hspace{-0.75em}
\includegraphics[width=0.245\textwidth]{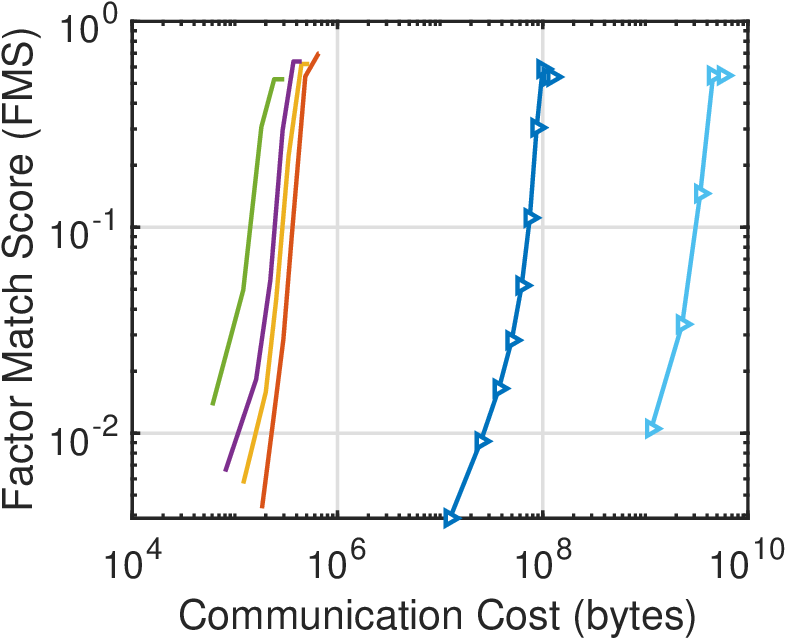}
\caption{Factor Match Scores (FMS) with respect to time and communication.}
\label{fig:fms}
\end{figure}

In order to validate the ability of extracting effective phenotypes using {\tt CiderTF}, we conduct a case study on MIMIC-III dataset. We evaluate the extracted phenotypes from both quantitative and qualitative perspectives. From the quantitative aspect, we use the Factor Match Score (FMS) \cite{acar2011scalable} to measure the similarity of the factor matrices of {\tt CiderTF} with the centralized baseline BrasCPD. FMS ranges from 0 to 1 with the best possible value of 1.

Fig. \ref{fig:fms} indicates that {\tt CiderTF} achieves the highest FMS of the decentralized methods as it gradually approaches 1. This means that {\tt CiderTF} can extract the factor matrices similar to its centralized counterpart BrasCPD. We can also observe that {\tt CiderTF} approaches the centralized factors with the least time and communication cost.

Furthermore, from the the qualitative perspective, we evaluate the quality of the phenotypes by patient subgroup identification ability. Following the precedent set in \cite{perros2017spartan}, we first identify the top three phenotypes according to the phenotype importance factor $\lambda_r={\left\|{\textbf{A}_{(1)}}(:,r)\right\|}_{F}{\left\|{\textbf{A}_{(2)}}(:,r)\right\|}_{F}\cdots{\left\|{\textbf{A}_{(D)}}(:,r)\right\|}_{F}$. We then group the patients by assigning each according to the largest value among the top 3 along the patient representation vector, and use tSNE to map the patient representation into two-dimensional space. Fig. \ref{tab:tsne} shows that {\tt CiderTF} ($\tau=8$) achieves comparable patient subgroup identification ability as the centralized baseline BrasCPD. When compared to the decentralized baselines with the same communication cost (since 1 epoch of D-PSGD and D-PSGDbras already incurs more communication cost, we show the result after 1 epoch), {\tt CiderTF} achieves better clustered subgroups, demonstrating that {\tt CiderTF} is able to better identify patient subgroups. In addition, the top 3 phenotypes extracted by {\tt CiderTF}, shown in table \ref{tab:pheno1}, are clinically meaningful and interpretable as annotated by a pulmonary and critical care physician.

\begin{table*}[h!]
\centering
\caption{tSNE visualization of the patient subgroup identification with the extracted phenotypes. Each point represents a patient which is colored according to the highest-valued coordinate in the patient representation vector among the top 3 phenotypes extracted based on the factor weights $\lambda_r={\left\|{\textbf{A}_{(1)}}(:,r)\right\|}_{F}{\left\|{\textbf{A}_{(2)}}(:,r)\right\|}_{F}\cdots{\left\|{\textbf{A}_{(D)}}(:,r)\right\|}_{F}$.}
 \begin{tabular}{ c c c c c c }
  \textbf{BrasCPD} & \makecell[c]{\textbf{D-PSGD}\\\textbf{1 epoch: $\mathbf{10^9}$}} & \makecell[c]{\textbf{D-PSGDbras}\\\textbf{1 epoch: $\mathbf{10^7}$}} & \makecell[c]{\textbf{SPARQ-SGD}\\ \textbf{$\mathbf{3.2\times10^5}$}} & \makecell[c]{\textbf{{\tt CiderTF} ($\tau=8$)}\\ \textbf{$\mathbf{10^5}$}} & \makecell[c]{\textbf{{\tt CiderTF} ($\tau=8$)}\\ \textbf{Total: $\mathbf{3.2\times10^5}$}} \\
 \includegraphics[width=0.14\textwidth,trim={1.5cm 1.5cm 2cm 1cm},clip]{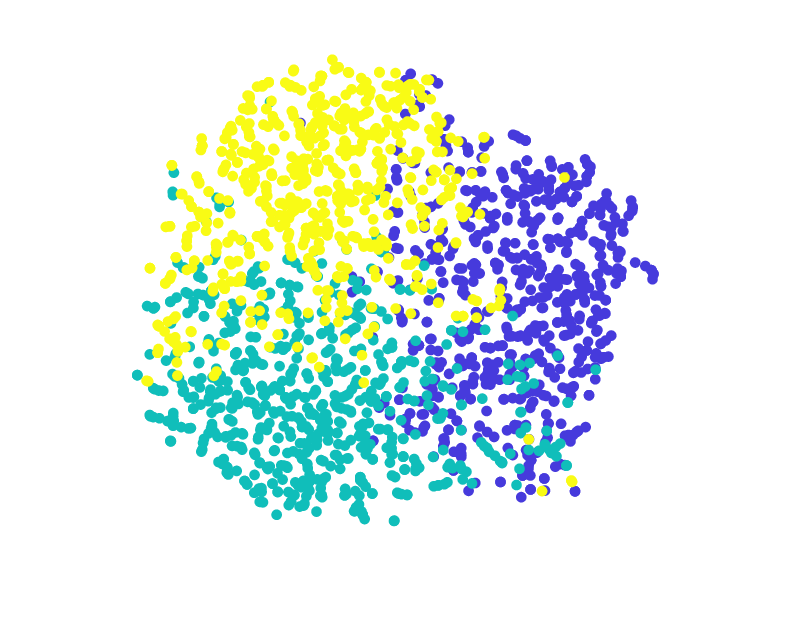}
  & 
  \includegraphics[width=0.14\textwidth,trim={1.5cm 1.5cm 2cm 1cm},clip]{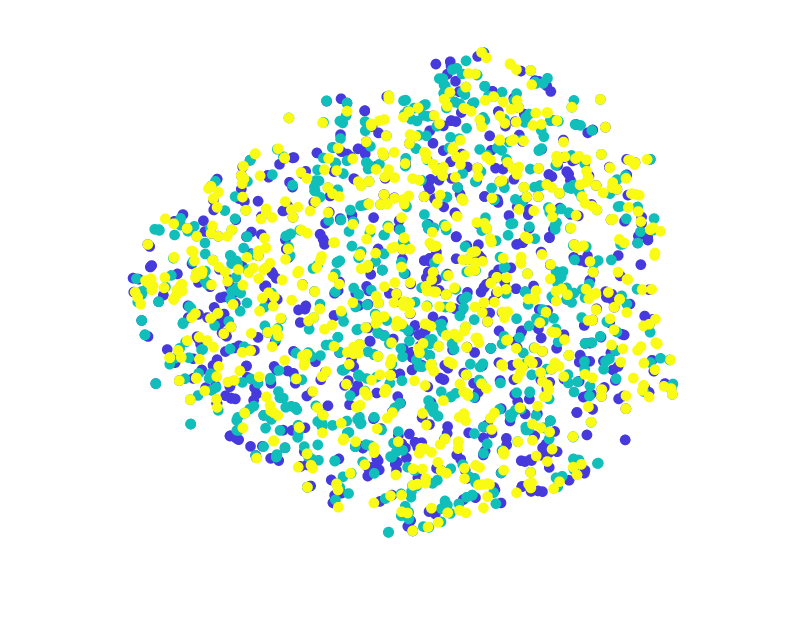}
  &
  \includegraphics[width=0.14\textwidth,trim={1.5cm 1.5cm 2cm 1cm},clip]{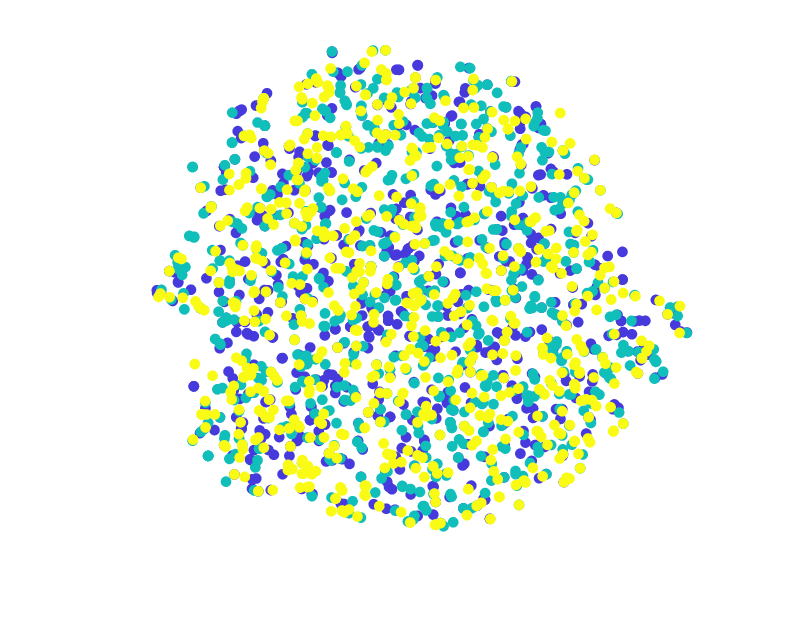}
  &
  \includegraphics[width=0.14\textwidth,trim={1.5cm 1.5cm 2cm 1cm},clip]{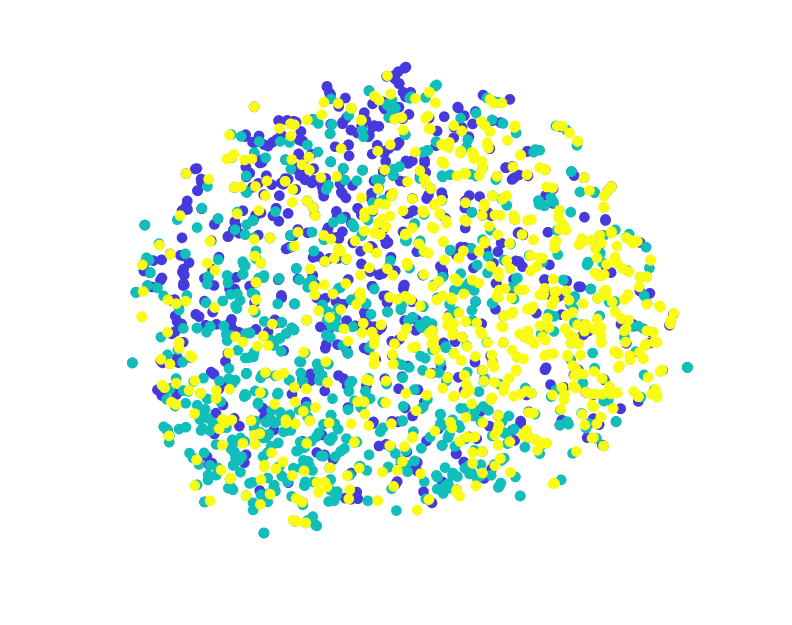}
  &
  \includegraphics[width=0.14\textwidth,trim={1.5cm 1.5cm 2cm 1cm},clip]{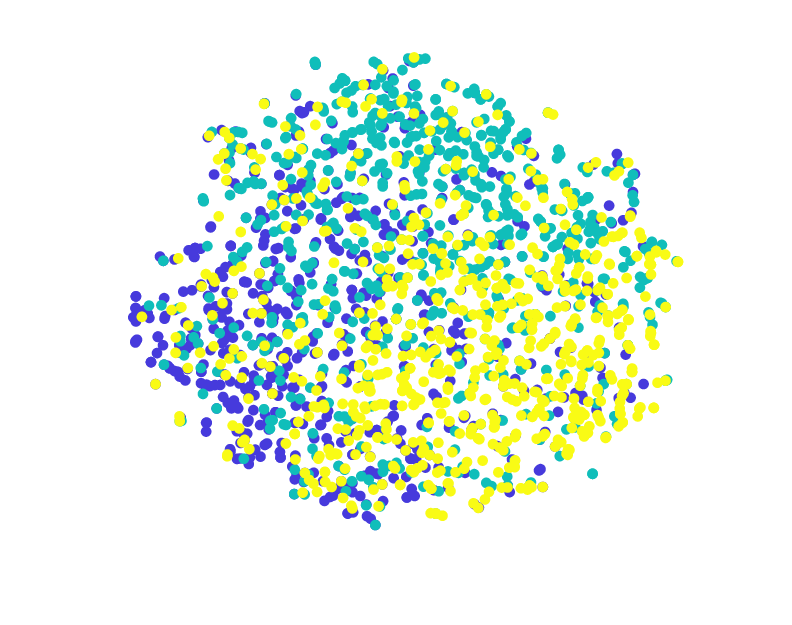}
  &
  \includegraphics[width=0.14\textwidth,trim={1.5cm 1.5cm 2cm 1cm},clip]{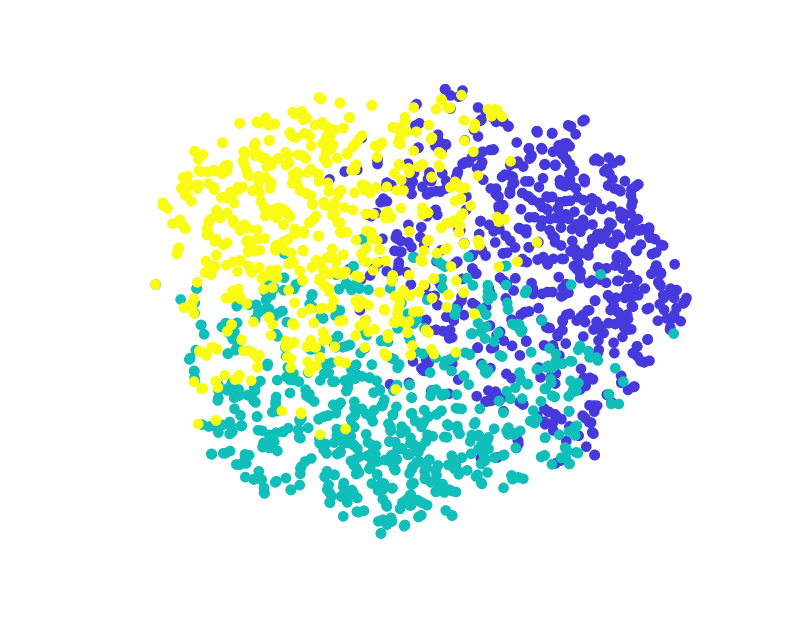}
\end{tabular}
\label{tab:tsne}
\end{table*}

\begin{table}[b]
\small
\setlength{\tabcolsep}{2pt}
\caption{phenptypes extracted by {\tt CiderTF} ($\tau=8$) on MIMIC-III data. Dx, Px, and Med indicate diagnoses, procedures, and medication.}
  \centering
  \begin{tabular}{p{0.7cm} l}
    \toprule
    \multicolumn{2}{l}{\textbf{P1: Acute myocardial infarction}}\\
    \midrule
    \multirow{2}*{Dx} & Other and unspecified angina pectoris\\
    ~& Coronary atherosclerosis of autologous vein bypass graft\\
    ~& Old myocardial infarction\\
    \midrule
    \multirow{2}*{Px} & (Aorto)coronary bypass of two coronary arteries\\
    ~& (Aorto)coronary bypass of three coronary arteries\\
    ~& Implant of pulsation balloon\\
    \midrule
    \multirow{2}*{Med} & Diltiazem Hydrochloride Extended-Release\\
    ~&Metoprolol succinate, Rosuvastatin Calcium\\
    ~&Valsartan/hydrochlorothiazide, Losartan Potassium\\
    \bottomrule
    \end{tabular}
\label{tab:pheno1}
\end{table}

\begin{table}[b]
\small
\setlength{\tabcolsep}{2pt}
  \centering
  \begin{tabular}{p{0.7cm} l}
    \toprule
    \multicolumn{2}{l}{\textbf{P2: Respiratory failure}}\\
    \midrule
    \multirow{2}*{Dx} & Acute respiratory failure, Hypoxemia, \\
    ~&Contusion of lung without mention of \\
    ~& \-\hspace{0.5cm}open wound into thorax\\
    ~&Disruption of internal operation (surgical) wound\\
    \midrule
    \multirow{2}*{Px} & Non-invasive mechanical ventilation\\
    ~&Continuous invasive mechanical ventilation for less than\\  ~& \-\hspace{0.5cm} 96 consecutive hours\\
    \midrule
    Med & Dextrose, Albuminar-25, Plasmanate\\
    \midrule
    \multicolumn{2}{l}{\textbf{P3: Intracranial hemorrhage or cerebral infarction}}\\
    \midrule
    \multirow{2}*{Dx} &Pure hypercholesterolemia, Subdural hemorrhage\\
    ~&Cerebral artery occlusion\\
    \midrule
    \multirow{2}*{Px} &Injection or infusion of thrombolytic agent\\
    ~&Control of hemorrhage\\
    \midrule
    Med &Ticagrelor, Atorvastatin Calcium\\
    \bottomrule
\end{tabular}
\label{tab:pheno2}
\end{table}

\section{Conclusion}
In this paper, we explore the communication efficient generalized tensor factorization under the decentralized optimization paradigm for computational phenotyping. We propose {\tt CiderTF}, which is the first decentralized generalized tensor factorization framework. It employs four levels of communication reduction with gradient compression, block randomization, and periodic communication combined with an event-driven communication strategy. Meanwhile, {\tt CiderTF} enjoys low computational and memory complexity due to the two levels of randomization: random fiber sampling and random block selection. Experiments show that {\tt CiderTF} preserves the quality of the extracted phenotypes and converges to similar points as the decentralized SGD baselines with theoretical guarantees. Future works including developing asynchronized communication and variance reduced techniques to the decentralized paradigm.

\section*{Acknowledgment}
This work was supported by the National Science Foundation under award IIS-\#1838200 and CNS-1952192, National Institute of Health (NIH) under award number R01LM013323, K01LM012924 and R01GM118609, CTSA Award UL1TR002378, and Cisco Research University Award \#2738379.

\bibliographystyle{IEEEtran}
\bibliography{gtf_network_ref}

\end{document}